\documentclass{bmvc2k}
\usepackage{amssymb}
\usepackage[table]{xcolor}
\usepackage{multirow}
\usepackage{graphicx}
\usepackage{epsfig}
\usepackage{amsmath}
\title{\\Emotion-Based Crowd Representation \\for Abnormality Detection}

\addauthor{Hamidreza Rabiee}{hr.rabiee@gmail.com}{1} 
\addauthor{Javad Haddadnia}{jhaddadnia@yahoo.com}{1}
\addauthor{Hossein Mousavi}{hossein.mousavi@iit.it}{2}
\addauthor{Moin Nabi}{moin.nabi@unitn.it}{3}
\addauthor{Vittorio Murino}{vittorio.murino@iit.it}{2}
\addauthor{Nicu Sebe}{niculae.sebe@unitn.it}{3}

\addinstitution{Hakim Sabzevari University, Iran}
\addinstitution{PAVIS, Istituto Italiano di \\Tecnologia, Genova, Italy}
\addinstitution{DISI, University of Trento, Italy}
\runninghead{Emotion-Based Crowd Representation for Abnormality Detection}{}

\begin{document}
\maketitle
\begin{abstract}
In crowd behavior understanding, a model of crowd behavior need to be trained using the information extracted from video sequences. 
Since there is no ground-truth available in crowd datasets except the crowd behavior labels, most of the methods proposed so far are just based on low-level visual features. However, there is a huge semantic gap between low-level motion/appearance features and high-level concept of crowd behaviors. In this paper we propose an attribute-based strategy to alleviate this problem. While similar strategies have been recently adopted for object and action recognition, as far as we know, we are the first showing that the crowd emotions can be used as attributes for crowd behavior understanding. The main idea is to train a set of emotion-based classifiers, which can subsequently be used to represent the crowd motion. For this purpose, we collect a big dataset of video clips and provide them with both annotations of ``crowd behaviors'' and ``crowd emotions''. We show the results of the proposed method on our dataset, which demonstrate that the crowd emotions enable the construction of more descriptive models for crowd behaviors. We aim at publishing the dataset with the article, to be used as a benchmark for the communities.
\end{abstract}
	
\section{Introduction}
Crowded analysis gained so much popularity in the recent years for both academic purposes and industrial AI. This growing trend is mainly due to the increase of population growth rate and the need of more precise public monitoring systems. Such systems are shown to be effective to capture crowd dynamics for public environments design~\cite{moussaid2011simple}, simulate crowd behavior for games designing~\cite{silverman2006human}, and group activity~\cite{nabi2013temporal} and crowd monitoring for visual surveillance~\cite{su2013large,mehran2009abnormal,benabbas2011motion}.
A human expert observer seem to be capable of monitoring the scene for unusual events in real time and taking immediate reactions ~\cite{hospedales2012video} accordingly. However, psychological research shows that the ability to monitor concurrent signals is really limited also in humans~\cite{sulman2008effective}. In the extremely crowded scenes, in presence of multiple individuals doing different behaviors, monitoring is a critical issue even for a human observer.
In the last few years, the computer vision community has pushed on crowd behavior analysis and particularly has made a lot of progresses in crowd abnormality detection~\cite{mehran2009abnormal,mousavi2015crowd,mousavi2015analyzing,mousavi2015abnormality}. However lack of publicly available \emph{realistic} datasets (i.e., high density, various types of behaviors, etc.) led not to have a fair common test bed for researchers to compare their algorithms and fairly evaluate the strength of their methods in real scenarios.
Besides the data, most of the proposed approaches only employ low-level visual features (see in ~\cite{su2013large,mehran2009abnormal,roggen2011recognition,saxena2008crowd,rodriguez2011data}, e.g., motion and appearance) to build the abnormal behavior models in crowd. However, many behaviors can be hardly characterized by simply low-level features and need to be represented with a more semantic description. The huge semantic gap between low-level video pixels and high-level crowd behavior concepts make a barrier itself to realize the dream of having a deep understanding of the behaviors in the scene.
\begin{figure*}
\begin{center}
\includegraphics[width=4.5in]{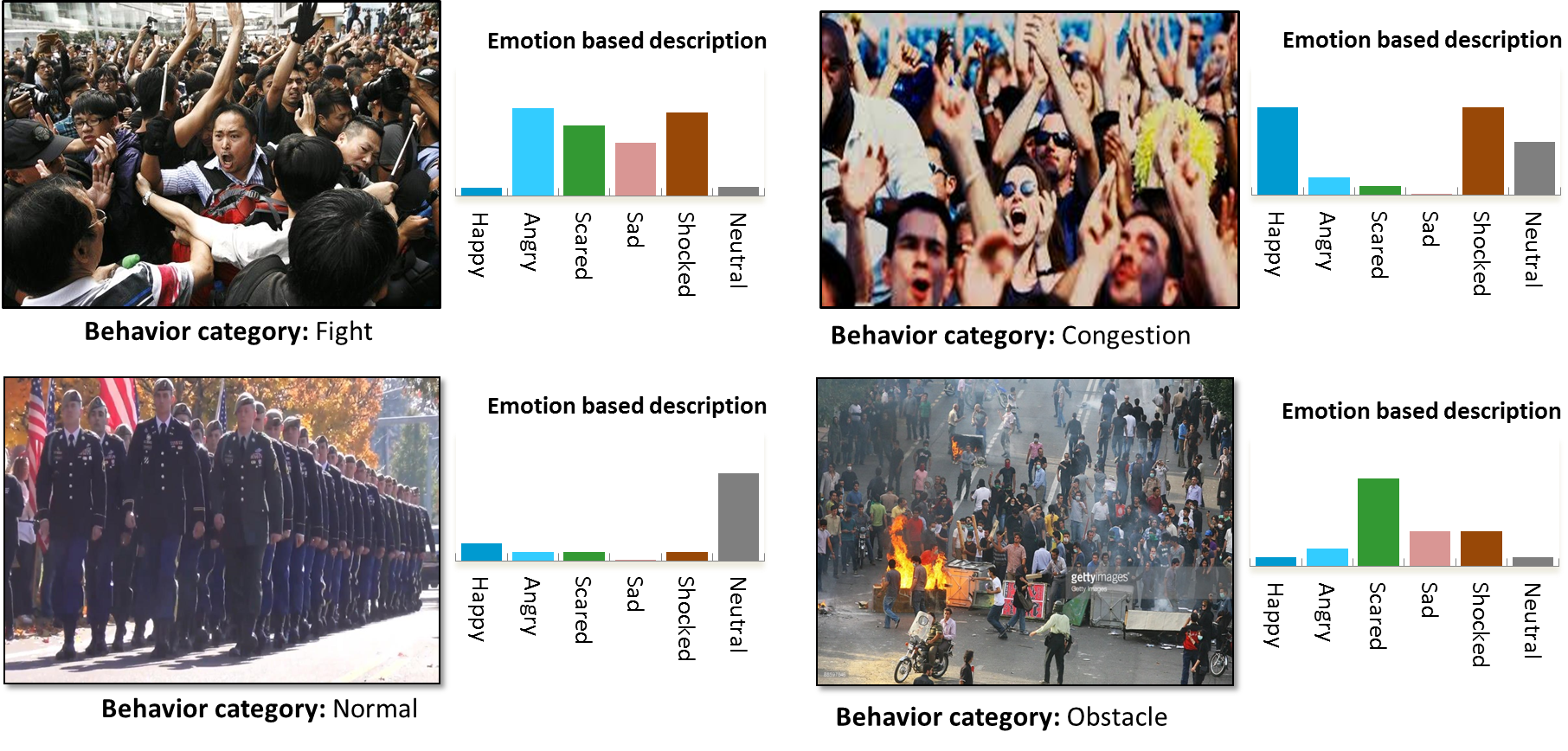}
\caption{Schematic illustration of the emotion-based crowd representation.}
\end{center}
\label{fig:panicfight}
\end{figure*}	
For example, crowd behaviors which are very similar visually, are appeared as entirely different in terms of being normal/abnormal, by only considering the \emph{emotion} of the individuals participating them (See Fig. \ref{fig:panicfight}).
Despite the relatively huge literature on emotion recognition for face~\cite{pantic2000automatic,bassili1979emotion,busso2004analysis} and posture~\cite{coulson2004attributing,mota2003automated}, there is just a few works which target emotion recognition from crowd motion~\cite{baig2015perception,baig2014crowd,mchugh2010perceiving}, and to the best of knowledge, there is no work which targets emotion recognition and abnormality behavior detection in a unified framework.
Inspired by the recent works on attribute-based representation in object/action recognition literature~\cite{liu2011recognizing,lampert2009learning,farhadi2010attribute}, we believe that human emotions can be used to represent a crowd motion, thus helping for crowd behavior understanding. In other words, to come up with a deep understanding of crowd behavior, the human emotions can be deployed to narrow down the semantic gap between low-level motion information and high-level behavior semantics.
Having this idea in mind, we create a dataset which includes both annotations of \emph{crowd behavior} and \emph{crowd emotion}. Such dataset opens up avenues for both tasks of abnormality detection and emotion recognition as well as analyzing the correlations of these two tasks. Our dataset consists of a big set of video clips annotated with crowd behavior labels (e.g., ``panic'', ``fight'', ``congestion'', etc.) and crowd emotion labels (e.g., ``happy'', ``excited'', ``angry'', etc.). 
We, first, evaluated a set of baseline methods on both tasks, showing that our dataset can be effectively used as a benchmark in both communities. We, then, analyzed the importance of crowd emotion information for crowd behavior understanding, by predicting the crowd behavior label using ground-truth emotion information.
And finally, we introduced a emotion-based crowd representation,
which provides us to exploit emotion information even without accessing to any ground-truth emotion information in testing time.
Our major contributions in this paper are: \textbf{(i)} Introducing a novel dataset of crowd with both behavior and emotions annotations. \textbf{(ii)} Showing the tight link between the two tasks of crowd behavior understanding and emotion recognition. \textbf{(iii)} Presenting a method which exploits jointly the complimentary information of these two task, outperforming all baselines of both tasks significantly.
The rest of the paper is organized as follows: a short review on previous datasets and the specification of our proposed dataset is reported in Sec. 2; the idea of emotion-based crowd representation for abnormal behavior detection is introduced in Sec. 3. In Sec. 4 we explain the considered baselines and the experiments regarding our introduced emotion-based approaches and discussing on the obtained results. Finally, in Sec. 5, we conclude the paper with briefly discussing other applications worth investigating and promoting further research on this new dataset.
\begin{table*}
	\begin{center}
	\renewcommand{\arraystretch}{1}
	\label{tab:datasetscompare}
	\scalebox{0.6}{
	\begin{tabular}{p{2.9cm}|lp{1.5cm}llllll}
		\noalign{\hrule height 1.3pt}
		Dataset   & UMN ~\cite{mehran2009abnormal} & UCSD~\cite{mahadevan2010anomaly} & CUHK~\cite{wang2009unsupervised} & PETS2009~\cite{ferryman2010pets2010} & ViF ~\cite{hassner2012violent} & Rodriguez's ~\cite{rodriguez2011data} & UCF ~\cite{solmaz2012identifying}& Our Dataset  \\
		\noalign{\hrule height 1.5pt}
		Number of samples  & 11 seq  & 98 seq  & 2 seq  & 59 seq & 246 seq  & 520 seq  & 61 seq & 31 seq\\
		\hline
		Annotation level  & frame  & frame/pixel & video & frame & video & video & video & frame\\
		\hline
		Density	  & semi   & semi & semi & semi & dense & dense  & dense & dense\\
		\hline
		Type of scenarios  & panic  & abnormal objects & traffic & panic & fight & pedestrians  & crowd & multi-category\\
		\hline
		Indoor/Outdoor & both & outdoor &  outdoor & outdoor & outdoor & outdoor & outdoor & outdoor\\
		\hline
		Meta-data & no & no & no & no & no & no & no &  crowd emotion\\
		
		\noalign{\hrule height 1.5pt}
	
		\end{tabular}
	}
\end{center}
\caption{Datasets for crowd behavior analysis}
\end{table*}
\section{The crowd anomaly detection dataset}
In this section, after a brief review on the well-known datasets for the task of crowd behavior analysis, we introduce our dataset in details.
\subsection{Previous Datasets}
Despite the significant demand for understanding crowd behaviors, 
there is still a significant gap between accuracy and efficiency of typical behavior recognition methods in research labs and the real world. The most important reason is that the majority of proposed algorithms are experimented on non-standard datasets having only a small number of sequences taken under controlled circumstances with limited behavior classes. \\
Among all the existing datasets, we select a subset of most cited ones and analyze the characteristics of them. The datasets that we select are: 
UMN ~\cite{mehran2009abnormal}, UCSD ~\cite{mahadevan2010anomaly}, CUHK ~\cite{wang2009unsupervised}, PETS ~\cite{ferryman2010pets2010} , Violent-Flows(ViF)  ~\cite{hassner2012violent}, Rodrigues's ~\cite{rodriguez2011data} and UCF  ~\cite{solmaz2012identifying}.
We also select a set of criteria which provide us the possibility of being able to compare the datasets according to them.
The criteria are including: \emph{number of samples, annotation level, crowd density, type of scenarios, indoor/outdoor, meta-data}.\\
\emph{Number of samples} is an important characteristics of a dataset, because having sufficient number of recorded videos can be so helpful not only for training with more samples, but also for evaluation. \emph{Annotation level} is another important criteria of the dataset. It can be specified as pixel-level, frame-level and video-level, which technically reflects the richness of a dataset in terms of annotation. \emph{Density} is another important factor of the crowd which should be considered. crowd scenes results in more occlusion, clutter scenarios which harder to detect different type of behaviors.
 \emph{Type of scenarios} is another critical characteristics of a dataset, reflects the type of events happenings in all the video sequences. Datasets with higher type of scenarios are more challenging, since the proposed algorithms should work on a larger variations of conditions (i.e., real world setup).
The \emph{Indoor/Outdoor} criteria is a about the location that the video sequences have been recorded and it has a peculiar effect on illumination conditions, background clutters, occlusions, etc.\\
Last but not least, \emph{Meta-data} is another important feature of a dataset, which we insist on it in this paper. It is also one of the features which make our dataset unique and provide the possibility for researchers to move toward higher-level interpretations of the video sequences. In our dataset, we specifically, introduced ``crowd emotion'' as extra annotation.
In table \ref{tab:datasetscompare} we describe all aforementioned crowd behavior datasets in terms of the explained features. A common demerit lies in all of them is lack of any extra annotation so that they all potentially only can be relied on low-level features to discriminate types of behavior classes. Lack of diverse subjects and scenarios, low volume crowd density and limited number of video sequences are other limitations of the previous datasets.
\subsection{The Proposed Dataset}
The presented dataset is consists of 31 video sequences in total or as about 44,000 normal and abnormal video clips. The videos were recorded as 30 frames per second using a fixed video camera elevated at a height, overlooking individual walkways and the video resolution is $554\times235$.
The crowd density in the crowd was changeable, ranging from sparse to very crowded. In addition to normal and abnormal behavior scenarios, a few scenarios including abnormal objects which can be regarded as threats to the crowd are also recorded make them more realistic. ``Motorcycle crossing the crowded scene'', ``a suspicious backpack left by an individual in the crowd'', ``a motorcycle which is left between many people'', etc. are some examples of such scenarios.
In proposed dataset, we have represented five typical types of crowd behaviors. Each scenario topology was sketched in harmony with circumstances usually met in crowding issues. They accord with a transition on a flow of individuals in a free environment (neutral), a crowded scene containing abnormal objects (obstacles), evacuation of individuals from the scene (panic), physical altercation between individuals (Fight), group of people gathering together (congestion). For each behavior type we recorded videos from two field of views with different crowd densities changing from sparse to very crowded.
All the videos in dataset starts with normal behavior frames and ends with abnormal ones. For emotion annotation we follow the standard definition of emotion in psychology which ``a feeling evoked by environmental stimuli or by internal body states'' ~\cite{bower2014emotional}. This can modulate human behavior in terms of actions in the environment or changes in the internal status of the body. There is no crowd abnormal behavior dataset available with emotion annotation in computer vision literature, so we following the datasets for face and gesture, we defined six types of basic emotions in our dataset, including: ``Angry'', ``Happy'', ``Excited'', ``Scared'', ``Sad'', ``Neutral''.
Since the perception of each of aforementioned emotions can be so subjective, we ask different annotators to annotate separately and finally we conclude over them using majority voting. To insure consistency between annotators, we conducted an agreement study, finding that the overall agreement is 92\% with a Kappa value of 0.81 and the maximum inconsistency was between Happy with Excited (confused 4\% of the time). 
 As we expected, our observation confirms that emotion annotations can provide a more descriptive representation for crowd behavior and it is clear even from the annotations. Specifically, in different circumstances, emotions belonging to an individual or a crowd have a crucial effect on decision making process. Emotion information is much richer than pure low-level motion information of videos.
 The videos, ground-truth annotations, and baseline codes will be available publicly soon after publishing the paper. We believe this dataset can be used as benchmark of future researches in both abnormal behavior detection and emotion recognition tasks.
\begin{figure*}
	\begin{center}
		\includegraphics[width=0.5\linewidth]{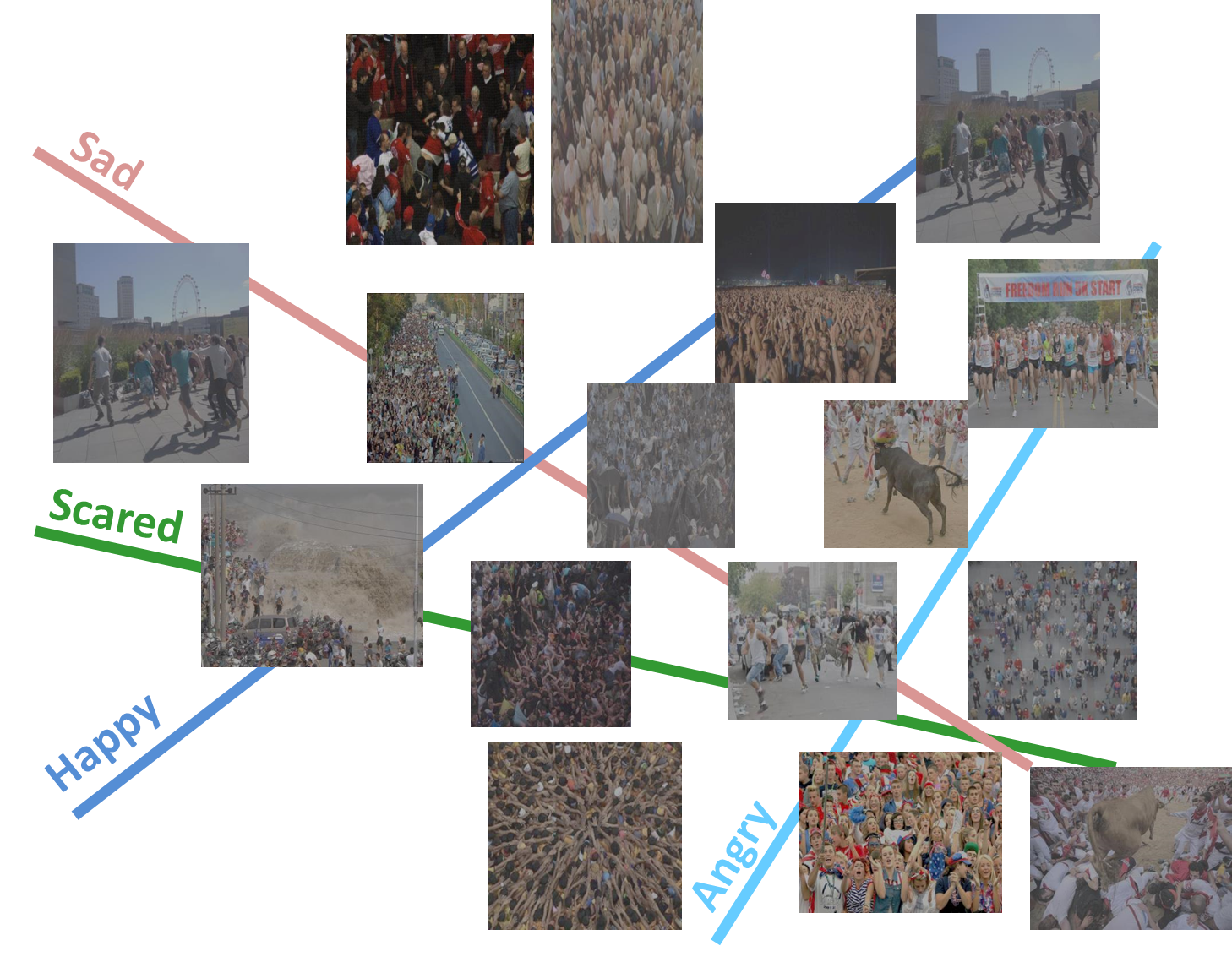}	

	\end{center}
		\caption{The emotion classifiers as mid-level representation for abnormal behaviors.}
	\label{fig:emotionbehavelabels}
\end{figure*}
\section{Emotion-Based Crowd Representation}
If the ground-truth emotion labels are available during both training and testing, we can simply consider them as part of the input data and solve a standard classification problem (See \emph{emotion-aware baseline} for evaluation, Sec.~\ref{seqemotionbased}). But things are not so straight forward when there is no ground-truth emotion label is available during testing. Here in this section we formally explain the latter.\\
Given set of $N$ video clips in the dataset $\{(x^{(n)},e^{(n)},y^{(n)})\}_{n=1}^N$, the goal is to learn 
a model employing emotion labels $e$ which be used to assign the abnormal behavior class label $y$ to an unseen test video clip $x$. In training phase, each example is represented as a tuple $(\textbf{f},\textbf{e},y)$ where $\textbf{f}\in\mathbb{F}^d$ is the $d$-dimensional low-level feature extracted from video clip $x$ (See Fig.~\ref{fig:emotionbehavelabels}). The abnormal behavior class label of the image is represented by $y\in\mathbb{Y}$. The crowd emotion of the video clip $x$ are denoted by a $K$-dimensional vector  $\textbf{e} = (e_1, e_2, ..., e_K)$, where $e_k\in\mathbb{E}_k$ $(k = 1, 2, ..., K)$ indicates the $k$-th emotion of the video clip.  For example, if the $k$-th emotion attribute is ``Angry'', we will have $\mathbb{E}_k = \{0, 1\}$, where $e_k$ = 1 means the crowd is ``Angry'', while $e_k$ = 0 means it is not. Since our datasets is designed to be applied also for standard multi-label emotion recognition setup, here, we describe each video clip with a binary-valued emotion attribute with a single non-zero value, i.e. $\mathbb{E}_k = \{0, 1\}$ $(e = 1, 2, ..., K)$, s.t. $\|\textbf{e}\|_0=1$. But we emphasize that our proposed method is not limited to binary-valued attributes with single emotion and simply can be extended to multi-emotion and continuous valued (fuzzy) attributes. Discarding the emotion information we can simply train a classifier $ \mathcal{C}\colon\mathbb{F}^d\rightarrow\mathbb{Y}$, that maps the feature vector $f$ to a behavior class label $y$ (See \emph{low-level visual feature baseline} for evaluation, Sec.~\ref{seqlowlevel1}). However, by employing emotion as an mid-level attribute to represent the crowd motion, the classifier $C$ is decomposed into:
\begin{equation} 
	\label{eq1}
	\begin{split}
		\mathcal{H} & =\mathcal{B}(\mathcal{E}(f))  \\&
		\mathcal{E}\colon\mathbb{F}^d\rightarrow\mathbb{E}_k \ and \  \mathcal{B}\colon\mathbb{E}_k\rightarrow\mathbb{Y}  
	\end{split}
\end{equation}

where $\mathcal{E}$ includes $K$ individual emotion classifiers $\{\mathcal{C}_{e_i}(f)\}_{i=1}^n$, and each classifier $\mathcal{C}_{e_i}$maps $f$ to the corresponding $i$-th emotion of $\mathbb{E}^n$,  $\mathcal{B}$ maps an emotion attribute  $\textbf{e}\in\mathbb{E}^n$ to a behavior class label  $ \textbf{y}\in\mathbb{Y}$. 
The emotion classifiers are learned during training using the emotion annotations provided by our dataset. Particularly, the classifier $ \mathcal{C}_{e_i}(f)$ is a binary linear SVM trained by labeling the examples of all behavior classes whose emotion value $e_i=1$ as positive examples and others as negative.
Assuming there is no emotion ground-truth information is available in test time, we represent each video clip $x$ by $\phi(x) \in \mathbb{E}_k$: 
\begin{equation} \label{eq2}
	\phi(x)  = [s_1(x),s_2(x),...,s_K(x)]
\end{equation}
where $s_k(x)$ is the confidence score of $k$-th emotion classifier $ \mathcal{C}_{e_k}$ in $\mathcal{E}$. This \emph{emotion-based crowd representation} vector has an entry for each emotion type reflects the degree to which a particular emotion is present in a video clip (See \emph{emotion-based crowd representation experiments}, Sec. ~\ref{seqemotionbased}). The mapping $\mathcal{B}$ is finally obtained by training a multi-class linear SVM for behavior classes on emotion-based crowd representation vectors. \\

\noindent\textbf{Latent Emotion:}
As aforementioned, we select crowd emotions as attributes, which are discriminative and yet able to extract the intra-class changing of each behavior. Note that intra-class changing may cause video clips to be associated with different sets of emotion information, in spite of belonging to same behavior class. For instance, the behavior class \textit{Congestion} in some video clips of a dataset might have the \textit{Angry} emotion attribute, while in other dataset it may contains \textit{happy} emotion attribute.(see Fig. \ref{fig:panicfight}). To address this problem, we treat emotion attributes as latent variables and learn the model using the latent SVM ~\cite{felzenszwalb2008discriminatively, wang2009max}. We aim at learning a classifier $f_W$ to predict the behavior class of an unknown video clip $x$ during testing. Specifically, a linear model is defined as:

\begin{equation}\label{equation3}
	W^T \Psi (x,y,e)=W_x \psi_1 (x) +\sum_{l\in\mathbb{E}} W^T_{e_l} \psi_2 (x,e_l) + \sum_{l,m\in\mathbb{E}} W^T_{e_l, e_m} \psi_3 (e_l,e_m)
\end{equation}
where parameter vector $W$ is $W=\{W_x;W_{e_l};W_{e_l,e_m}\}$, and $\mathbb{E}$ is an emotion attribute set. The first term in Eq.~\ref{equation3} provides the score measuring how well the raw feature $\psi_1 (x)$ of a video clip matches the template $W_x$ which is a set of coefficients learned from the raw features $x$. The second term in Eq.~\ref{equation3} provides the score of a specific emotion, and is used to indicate the presence of a emotion in the video clip $x$.
The initial value of $e_l$ is inherited from the behavior label during training, and is given by a pre-trained emotion classifier in the testing phase (see Sec 3). The third term defined to capture the co-occurrence of the pairs of emotions. Specifically, the feature vector $\psi_3$ of a pair $e_l$ and $e_m$ is a $\mathbb{E}\times\mathbb{E}$ dimensional indicator for the pair configurations and the associated $W^T_{e_l, e_m}$ contains the weights of all configurations.
From a set of training instances, the model vector $W$ is learned by solving the following formulation as learning objective function:
\begin{equation}\label{equation4}
	W^*=\min_{W} \lambda||W||^2+\sum_{j=1}^{n} \max(0,1-y_j.f_w(x_j))
\end{equation}
where $\lambda$ is the trade-off parameter controlling the amount of regularization, and the second term performs a soft-margin. 
Since the objective function in Eq.~\ref{equation4} is semi-convex, a local optimum can be obtained by the coordinate descent ~\cite{felzenszwalb2008discriminatively}. 
In our current implementation, each emotion has two statuses \{0\} and \{1\} and belief propagation ~\cite{felzenszwalb2008discriminatively} is employed to find the best emotion configurations(See \emph{latent-emotion crowd representation experiments}, Sec. ~\ref{seqemotionbased}).
 
 \section{Experiments}\label{exp}
In this section, we first explain the baseline methods we used to extract low-level visual features from our dataset. We, then, explain the experiments regarding emotion-based crowd representation for crowd behavior understanding. Note that during the experiments we fixed the evaluation protocol during all the experiments. We divide the train and test data in a leave-one-sequence-out fashion. More specifically, for 31 times (equal to number of video sequences) we leave one video clip of a sequence out for test and train on all the remaining 30. As the evaluation measure we used the average accuracy both in tables and confusion matrices. Note that all confusion matrices are based on dense trajectory features.

\subsection{Baseline Methods}\label{seqlowlevel1} 
\noindent\textbf{Low-level Visual Feature Baseline:}\label{baseline1}
As low-level features, we exploited the well-known dense trajectories ~\cite{wang2011action,wang2012abnormal} by extracting them for each video clips using the code provided by~\cite{wang2011action}. For this purpose, we computed state-of-the-art feature descriptors, namely histogram of oriented gradients (HOG)~\cite{dalal2005histograms}, histogram of optical flow (HOF)~\cite{laptev2008learning}, motion boundary histogram (MBH)~\cite{dalal2006human} and dense trajectories ~\cite{wang2011action} within space-time patches to leverage the motion information in dense trajectories. The size of the patch is $32\times 32$ pixels and 15 frames.\\ 
We use the bag-of-words representation for each clip to create a low-level visual feature using the extracted feature descriptors. In particular, we extract a codebook for each descriptor (HOG, HOF, MBH, and Trajectories) by fixing the number of visual words to $d$=1000. For the sake of time, we just did clustering on a subset of randomly selected training features using k-means. Descriptors are allocated to their closest vocabulary word using Euclidean distance. The extracted histograms of visual words finally used as a video descriptor. For classification of videos we use a standard one-vs-all multi-class SVM classifier. We separately evaluate HOG, HOF, MBH,Trajectory and Dense trajectory low-level feature descriptors by ground-truth label information of the behavior and the average accuracy of each is presented in first column of table \ref{tab:accuracylabel}. As can be seen, dense trajectory feature achieved 38.71 \% accuracy in crowd abnormality detection and has better performance comparing with four other feature descriptors.
Fig. \ref{fig:conf_all} (b) shows the performance comparison between varied combinations of different types of behavior categories in confusion matrix based on dense trajectory features. As you can see, the "\textit{Panic}" category has the best result of 74.82 \% compares to other behavior classes, probably due to solving a simpler task. The most confusion of this category was with "\textit{fight}" which can be justified as the similarity of motion patterns in these two categories (very sharp movements).
\begin{table}
	\begin{center}
		\scalebox{1}{
			\begin{tabular}{cccc}
				\noalign{\hrule height 1.5pt}
				& \multicolumn{3}{c}{Our dataset}\\
				{} & low-level visual feature  & emotion-aware & emotion-based\\
				\hline
				
				Dense trajectory & 38.71 & 83.79 & 43.64 \\
				
				\hline
				\noalign{\hrule height 1.5pt}
			\end{tabular}
		}
	\end{center}
	\caption{Comparison of dense trajectory descriptor on low-level visual features, emotion-aware and emotion-based categories in our dataset. We report average accuracy over all classes for our dataset}
	\label{tab:accuracylabel1}
\end{table}

\begin{table}
	\begin{center}
	\scalebox{1}{		
	\begin{tabular}{cccc}				
				\noalign{\hrule height 1.5pt}
				& \multicolumn{3}{c}{Our dataset}\\
				{} & low-level visual feature  & emotion- based & latent-emotion\\
				\hline
				Trajectory & 35.30  & 40.05 & 40.04\\
				HOG        & 38.80  & 38.77 & 42.18 \\ 
				HOF        & 37.69 & 41.50 & 41.51 \\
				MBH        & 38.53 & 42.72 & 42.92 \\
				\hline
				Dense Trajectory & \textbf{38.71}  & \textbf{43.64} & \textbf{43.90} \\ 
				\noalign{\hrule height 1.5pt}
			\end{tabular}
		}
	\end{center}
	\caption{Comparison of different feature descriptors (Trajectory, HOG, HOF, MBH and Dense Trajectory) on  low-level visual feature, emotion-based and latent-emotion categories in our dataset. We report average accuracy over all classes for our dataset.}
	\label{tab:accuracylabel}
\end{table}
\subsection{Emotion-based representation experiments}\label{seqemotionbased}
In this part we explain the experiments regarding our emotion-based proposed methods. In the first experiment, we assume to have access to emotion labels both in testing and training time and in the second experiment, instead, we have access to emotion labels only in training time.\\ 
\noindent\textbf{Emotion-aware Baseline:}
In this experiment, we use the ground-truth label information of the crowd emotion for feature construction. For this purpose, we first simply create a 6 dimension binary feature vector for all test and train data. As an example, since we have 6 emotion classes, namely "angry","happy","excited","scared","sad" and "neutral" respectively, a feature vector related to a video clip with emotion class "happy" is represented as \{0,1,0,0,0,0\} $\in\mathbb{E}^6$. Considering constructed features, we train a multi-class SVM classifier using the abnormal behavior labels.
In testing time we evaluate the trained classifier with test examples. In second column of table \ref{tab:accuracylabel1} we report the average accuracy over all behavior categories.
Such significant margins suggests that having a precise emotion recognition method can be so helpful for crowd behavior understanding. Inspired by this results, in following experiments we employing the emotion as mid-level representation for crowd motion. \\
\noindent\textbf{Emotion-based Crowd Representation Experiment:}
In this part, We first used the ground-truth label information of the emotion to separately evaluate aforementioned low-level feature descriptors. Fig. \ref{fig:conf_all} (c) shows the performance comparison between diverse combinations of different types of emotion categories in confusion matrix based on dense trajectory features with average accuracy of 34.13 \%. Based on the results reported in the confusion matrix Fig. \ref{fig:conf_all} (c), although the results are not so good, but it still can be used for abnormality behavior detection.
In this part we assume that there is no emotion label  available for test data, so we learn a set of binary linear SVMs using emotion labels of training data. We call this classifier as emotion classifiers $ \mathcal{C}_{e_i}(f)$. The output of emotion classifiers is a vector in which each dimension shows confidence score of emotion prediction. We consider this vector as an emotion-based crowd representation vector for behavior classification. We extract this vector for all train and test images and then train a multi-class SVMs with behavior labels. This behavior classifier is finally evaluated on test data to report the final accuracy of behavior classification.\\
We applied this method separately to HOG, HOF, MBH, Trajectory and Dense trajectory low-level feature descriptors.The average accuracy results for each of which is presented in second column of table. \ref{tab:accuracylabel}. Dense trajectory feature achieved the best precision with 43.64 \% among the other low level features. In compare with two other baselines, our method has highest accuracy and increase it by almost 7 \%.
Also in confusion matrix in Fig. \ref{fig:conf_all} (a), the best detection result belongs to "\textit{Panic}" behavior class with 71.87 \% and the most conflict to this class belongs to "\textit{fight}" behavior category with 11.88 \%. On the other hand, the worst detection result belongs to "congestion" behavior class with the most conflict of 21.92 \% to "\textit{panic}" behavior class. This results are in line with the average accuracies that we have for emotion based classifiers and emotion-aware baseline. This results supports that by having a better emotion recognition classifiers and more precise emotion labels we can boost the performance.\\
\noindent\textbf{Latent-Emotion Crowd Representation Experiment:}
Finally, we treated emotion labels as latent variables and learn the model using the latent SVM. In third column of table \ref{tab:accuracylabel}, the results are presented where 43.9 \% belongs to this experiment. It suggests that by using crowd emotion as mid-level representation we can obtain more accurate detection of crowd behavior classes. 
\begin{figure*}
	\begin{tabular}{cc}
		\newcommand{\cb}[1]{{\cellcolor{black! #1 }$ #1 \%$}}
\newcommand{\cw}[1]{{\cellcolor{black! #1 }$ \color{white} #1 \%$}}

	\scalebox{0.55}{
	\begin{tabular}{c | c c c c c c c c c c c c}
		\multicolumn{1}{c}{} & & \multicolumn{5}{c}{Prediction}& \\ \cline{3-8}
		 \multicolumn{1}{c}{} \\ & & \rotatebox[origin=c]{0}{Panic} & \rotatebox[origin=c]{0}{Fight} & \rotatebox[origin=c]{0}{Congestion} & \rotatebox[origin=c]{0}{Obstacle} & \rotatebox[origin=c]{0}{Neutral}\\
		\multirow{6}{*}{\rotatebox[origin=c]{90}{Truth}}
		
		 &\rotatebox[origin=c]{45}{Panic}	& \cw{71.87}  & \cb{11.88}	& \cb{7.49}	& \cb{4.64} 	& \cb{4.19} 	\\ 
		 &\rotatebox[origin=c]{45}{Fight}	& \cb{21.72}	& \cw{34.37} & \cb{13.24}  & \cb{18.76}	& \cb{11.91} 	\\ 
		 &\rotatebox[origin=c]{45}{Cong.} &\cb{21.92}  & \cb{18.98}	& \cw{30.66} & \cb{18.69}  & \cb{9.75}  \\ 
		 &\rotatebox[origin=c]{45}{Obstacle} &\cb{11.01}	& \cb{20.11}	& \cb{13.86}	& \cw{33.19} & \cb{21.83}  \\ 
		 &\rotatebox[origin=c]{45}{Neutral}	& \cb{10.11}	& \cb{12.67}	& \cb{8.46}	& \cb{20.65} & \cw{48.11} \\ 	 
	\end{tabular}}

%
 &\newcommand{\cb}[1]{{\cellcolor{black! #1 }$ #1 \%$}}
\newcommand{\cw}[1]{{\cellcolor{black! #1 }$ \color{white} #1 \%$}}

\scalebox{0.55}{
	\begin{tabular}{c | c c c c c c c c c c c c}
		\multicolumn{1}{c}{} & & \multicolumn{5}{c}{Prediction}& \\ \cline{3-8}
		\multicolumn{1}{c}{} \\& & \rotatebox[origin=c]{0}{Panic} & \rotatebox[origin=c]{0}{Fight} & \rotatebox[origin=c]{0}{Congestion} & \rotatebox[origin=c]{0}{Obstacle} & \rotatebox[origin=c]{0}{Neutral}\\
		\multirow{6}{*}{\rotatebox[origin=c]{90}{Truth}}
		
		 &\rotatebox[origin=c]{45}{Panic}	& \cw{74.82}  & \cb{15.18}	& \cb{5.64}	& \cb{3.39} 	& \cb{0.97} 	\\ 
		&\rotatebox[origin=c]{45}{Fight}	& \cb{24.48}	& \cw{30.47} & \cb{17.18}  & \cb{18.24}	& \cb{9.63} 	\\ 
		&\rotatebox[origin=c]{45}{Cong.} & \cb{32.17}  & \cb{18.11}	& \cw{23.43} & \cb{18.91}  & \cb{7.38}  \\ 
		&\rotatebox[origin=c]{45}{Obstacle} & \cb{9.25}	& \cb{25.54}	& \cb{19.02}	& \cw{27.94} & \cb{18.25}  \\ 
		&\rotatebox[origin=c]{45}{Neutral} & \cb{9.40}	& \cb{16.80}	& \cb{17.65}	& \cb{19.27} & \cw{36.88} \\ 	 
	\end{tabular}}

%

\\
		(a)&(b)\\
		\multicolumn{2}{c}{\newcommand{\cb}[1]{{\cellcolor{black! #1 }$ #1 \%$}}
\newcommand{\cw}[1]{{\cellcolor{black! #1 }$ \color{white} #1 \%$}}

\scalebox{.55}{
	\begin{tabular}{c | c c c c c c c c c c c c}
		\multicolumn{1}{c}{} & & \multicolumn{5}{c}{Prediction}& \\ \cline{3-8}
		\multicolumn{1}{c}{}\\ & & \rotatebox[origin=c]{0}{Angry	} & \rotatebox[origin=c]{0}{Happy} & \rotatebox[origin=c]{0}{Excited} & \rotatebox[origin=c]{0}{Scared} & \rotatebox[origin=c]{0}{Sad} & \rotatebox[origin=c]{0}{Neutral}\\
		\multirow{7}{*}{\rotatebox[origin=c]{90}{Truth}}
		
		&\rotatebox[origin=c]{45}{Angry}	   & \cw{25.42}  & \cb{15.40}	& \cb{16.12}	& \cb{26.45} 	& \cb{11.14} & \cb{5.47}  	\\ 
		&\rotatebox[origin=c]{45}{Happy}     & \cb{ 17.60}	& \cw{18.10} & \cb{23.92}  & \cb{15.05}	& \cb{19.06} & \cb{6.27} 	\\ 
		&\rotatebox[origin=c]{45}{Excited}   & \cb{20.39}  & \cb{11.90}	& \cw{32.22} & \cb{5.91}  & \cb{16.11} & \cb{13.47}   \\ 
		&\rotatebox[origin=c]{45}{Scared}  & \cb{14.02}	& \cb{10.22}	& \cb{6.58}	& \cw{65.92} & \cb{2.86} & \cb{0.40}  \\ 
		&\rotatebox[origin=c]{45}{Sad}	& \cb{26.92}	& \cb{6.75}	& \cb{6.31}	& \cb{27.66} & \cw{29.56} & \cb{2.80} \\ 
		&\rotatebox[origin=c]{45}{Neutral}	& \cb{9.59}	& \cb{17.88}	& \cb{17.51}	& \cb{7.54} & \cb{13.90} & \cb{33.58} \\
 
	\end{tabular}}

%
}\\
		\multicolumn{2}{c}{(c)}
		
	\end{tabular}
	\caption{(a): Confusion matrix for each emotion-based class. (b): Confusion matrix for each low-level visual feature  class. (c): Confusion matrix for six predefined emotion classes. }
	\label{fig:conf_all}
\end{figure*}
\section{Conclusions and Future Works}
In this paper, we have proposed a novel crowd dataset with both annotations of abnormal crowd behavior and crowd emotion. We believe this dataset not only can be used as a benchmark in computer vision community, but also can open up doors toward understanding the correlations between the two tasks of ``crowd behavior understanding'' and ``emotion recognition''.  As the second contribution we present a method which exploits jointly the complimentary information of these two task, outperforming all baselines of both tasks significantly. In particular, future work will be directed towards recognizing a novel abnormal behavior class with no training samples available, by manually defining the emotion-to-behavior mapping function.

\bibliography{egbib}

\end{document}